\pdfoutput=1
\documentclass[11pt,a4paper]{article}
\usepackage[hyperref]{acl2020}
\usepackage{times}
\usepackage{latexsym}

\usepackage{microtype}

\usepackage{amsmath}
\usepackage{amssymb}
\usepackage{arydshln}
\usepackage{booktabs}
\usepackage{tabularx}
\usepackage{tikz}
\usepackage{pgfplots}
\usepackage[normalem]{ulem}
\usepackage{xcolor}
\usepackage{dashbox}%

\newcommand\dashedph[1][H]{\setlength{\fboxsep}{0pt}\setlength{\dashlength}{2.2pt}\setlength{\dashdash}{1.1pt} \dbox{\phantom{#1}}}

\newcommand\mydots{%
  \mathinner{{\ldotp}{\ldotp}{\ldotp}}%
}

\pgfplotsset{compat=1.13}

\definecolor{c0}{cmyk}{1,0.3968,0,0.2588} 
\definecolor{c1}{cmyk}{0,0.6175,0.8848,0.1490} 
\definecolor{c2}{cmyk}{0.1127,0.6690,0,0.4431} 
\definecolor{c3}{cmyk}{0.6765,0.2017,0,0.0667} 
\definecolor{c4}{cmyk}{0.3081,0,0.7209,0.3255} 
\definecolor{c5}{cmyk}{0,0.8765,0.7099,0.3647}
\usetikzlibrary{calc,fit,positioning}
\DeclareMathOperator*{\argmax}{arg\,max}
\DeclareMathOperator*{\argmin}{arg\,min}

\makeatletter
\def\adl@drawiv#1#2#3{%
        \hskip.5\tabcolsep
        \xleaders#3{#2.5\@tempdimb #1{1}#2.5\@tempdimb}%
                #2\z@ plus1fil minus1fil\relax
        \hskip.5\tabcolsep}
\newcommand{\cdashlinelr}[1]{%
  \noalign{\vskip\aboverulesep
           \global\let\@dashdrawstore\adl@draw
           \global\let\adl@draw\adl@drawiv}
  \cdashline{#1}
  \noalign{\global\let\adl@draw\@dashdrawstore
           \vskip\belowrulesep}}
\makeatother
\aclfinalcopy % Uncomment this line for the final submission
\def\aclpaperid{309} %  Enter the acl Paper ID here

\title{\textsc{Bertram}:
  Improved Word Embeddings
  Have Big Impact on Contextualized Model Performance}

\author{Timo Schick \\ Sulzer GmbH \\ Munich, Germany \\ \texttt{timo.schick@sulzer.de} \And Hinrich Sch\"utze 
\\ Center for Information and Language Processing \\ LMU Munich, Germany \\ \texttt{inquiries@cislmu.org} \\
}

\date{}

\newcounter{notecounter}
\newcommand{\enotesoff}{\long\gdef\enote##1##2{}}
\newcommand{\enoteson}{\long\gdef\enote##1##2{{
\stepcounter{notecounter}
{\large\bf
\hspace{1cm}\arabic{notecounter} $<<<$ ##1: ##2
$>>>$\hspace{1cm}}}}}
\enoteson
\enotesoff

\long\def\eat#1{\ignorespaces}

\def\secref#1{\S\ref{sec:#1}}
\def\eqref#1{Eq.~\ref{eqn:#1}}

\begin{document}
\maketitle
\begin{abstract}

Pretraining deep language models has led to large performance
gains in NLP.
Despite this success, 
\citet{schick2019ota} recently showed that these models
struggle to understand rare words. For
static word embeddings, this problem has been
addressed by separately learning representations for
rare words. In this work, we transfer this idea to
pretrained language models:
%Most approaches for inducing word embeddings are
%simple bag-of-words models and are therefore
%not suitable  for deep language models.
%To overcome this problem,
We introduce \textsc{Bertram}, a powerful architecture
based on BERT that is capable of
inferring high-quality embeddings for rare words 
that are suitable as input representations for deep language models.
This is achieved by enabling the
surface form and contexts of a word
to interact with each other in a deep architecture.
Integrating \textsc{Bertram} into BERT
leads to large performance increases due to improved representations of rare and medium
frequency words on both a rare word probing
task and three downstream tasks.\footnote{Our implementation of \textsc{Bertram} is publicly available at \url{https://github.com/timoschick/bertram}.}
\end{abstract}

\section{Introduction}

As word embedding algorithms
\citep[e.g.][]{Mikolov2013} are known to struggle with rare
words, several techniques for improving their
representations have been proposed.
These approaches exploit either the contexts in which
rare words occur
\citep{lazaridou2017multimodal,herbelot2017high,khodak2018carte,liu-etal-2019-second},
their surface-form
\citep{luong2013better,bojanowski2016enriching,pinter2017mimicking},
or both \citep{schick2019attentive,schick2018learning,hautte2019bad}. However, all of this prior work
is designed for and evaluated on
\emph{uncontextualized} word embeddings.

Contextualized representations
obtained from pretrained deep language models
\citep[e.g.][]{peters2018deep,radford2018improving,devlin2018bert,liu2019roberta}
already handle rare
words implicitly using methods such as byte-pair encoding
\citep{SennrichHB15}, WordPiece embeddings
\citep{wu2016google} and  character-level CNNs
\citep{baevski2019cloze}.
Nevertheless, \citet{schick2019ota} recently showed that
BERT's \citep{devlin2018bert} performance on a rare word
probing task can be significantly  improved by explicitly
learning representations of rare words using Attentive
Mimicking (AM) \citep{schick2019attentive}. However, AM is limited in two important respects:
\begin{itemize}
\item For processing contexts, it uses a simple bag-of-words
model, making poor use of the available information. 
\item It combines form and context  in a shallow fashion,
   preventing both input signals from interacting in a complex manner.
\end{itemize}
These limitations apply not only to AM, but
to all previous work on obtaining representations
for rare words by leveraging form and context. While using
bag-of-words models is a reasonable choice for
static embeddings, which are often themselves
bag-of-words
\citep[e.g.][]{Mikolov2013,bojanowski2016enriching}, it
stands to reason that they are not the best choice to generate input representations for position-aware, deep language models.

To overcome these limitations, we introduce \textsc{Bertram}
(\textbf{BERT} fo\textbf{r} \textbf{A}ttentive
\textbf{M}imicking), a novel architecture for
learning rare word representations that combines a
pretrained BERT model with AM.
As shown in Figure~\ref{bertram-idea},
the learned rare word representations can
then be used as an improved input representation for another BERT model.
%Unlike previous approaches
%making use of language models \citep{liu-etal-2019-second},
%our approach integrates BERT in an end-to-end fashion and
%directly makes use of its hidden states. 
By giving
\textsc{Bertram} access to both surface form and contexts starting at the lowest layer, a
deep integration of both
input signals becomes possible.

Assessing the effectiveness of methods like \textsc{Bertram}
in a contextualized setting
is challenging: While most previous work on rare words was
evaluated on datasets explicitly focusing on rare words
\citep[e.g][]{luong2013better,herbelot2017high,khodak2018carte,liu-etal-2019-second},
these datasets are tailored to
uncontextualized embeddings and thus not suitable for
evaluating our model. Furthermore, rare words
are not well represented in
commonly used downstream task datasets.
We therefore introduce \emph{rarification}, a
procedure to automatically convert
evaluation datasets into ones for which rare words are guaranteed to be important. This is achieved by replacing task-relevant frequent words with rare synonyms obtained using semantic resources such as WordNet \citep{miller1995wordnet}.
We rarify
three common text (or text pair) classification
datasets: MNLI \citep{williams2018mnli}, AG's News
\citep{zhang2015character} and DBPedia
\citep{lehmann2015dbpedia}. \textsc{Bertram} outperforms previous
work on four English datasets by a large margin:
on the three rarified datasets and
on WNLaMPro 
\citep{schick2019ota}.

In summary, our contributions are as follows:
\begin{itemize}
\item We introduce \textsc{Bertram}, a model that integrates BERT into Attentive Mimicking, enabling a deep integration of surface-form and contexts and much better representations for rare words.
\item We devise rarification, a method that transforms evaluation
  datasets into ones for which rare words are guaranteed to be important.
\item We show that adding \textsc{Bertram} to BERT achieves a new state-of-the-art on WNLaMPro  \citep{schick2019ota} and beats all baselines on rarified AG's News, MNLI and DBPedia, resulting in an absolute improvement of up to 25\% over BERT.
\end{itemize}

\tikzset{
  tnode/.style={rectangle, inner sep=0.05cm, minimum height=2ex, text centered,text height=1.5ex, text depth=0.25ex},
  opnode/.style={draw, rectangle, rounded corners, minimum height=3ex, minimum width=4ex, text centered},
  arrow/.style={draw,->,>=stealth}
}
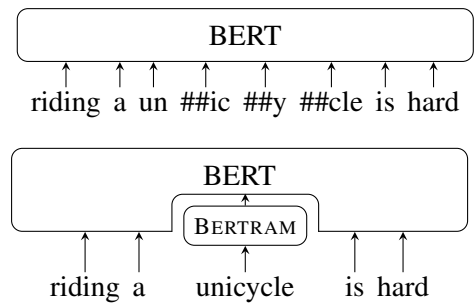
\begin{figure}
\centering
\begin{tikzpicture}

% context embedding nodes
\node[tnode](riding-b){riding};
\node[tnode,right=0.05cm of riding-b](a-b){a};
\node[tnode,right=0.05cm of a-b](un-b){un};
\node[tnode,right=0.05cm of un-b](ic-b){\#\#ic};
\node[tnode,right=0.05cm of ic-b](y-b){\#\#y};
\node[tnode,right=0.05cm of y-b](cle-b){\#\#cle};
\node[tnode,right=0.05cm of cle-b](is-b){is};
\node[tnode,right=0.05cm of is-b](hard-b){hard};

\node[fit=(riding-b)(hard-b)](text-b){};
\node[opnode, above=0.2cm of text-b, minimum width=6cm, minimum height=0.7cm](bert-b){BERT};

\node[tnode,below=1.9cm of text-b](unicycle-a){\phantom{unicycle}};
\node[tnode,left=0.55cm of unicycle-a](a-a){a};
\node[tnode, left=0.05cm of a-a](riding-a){riding};
\node[tnode,right=0.55cm of unicycle-a](is-a){is};
\node[tnode,right=0.05cm of is-a](hard-a){hard};

\node[fit=(riding-a)(hard-a)](text-a){};
\node[opnode, above=0.4cm of text-a, minimum width=6cm, minimum height=1.1cm](bert-a){};
\node[below,yshift=-0.1cm] at (bert-a.north) {BERT};

\node[opnode, above=0.35cm of unicycle-a, fill=white](bertram){\small\textsc{Bertram}};
\node[opnode, fit=(bertram), inner sep=0.15cm, fill=white](bertram-border){};
\node[fill=white, below=0cm of bert-a, minimum width=2.2cm, minimum height=0.6cm]{};
\node[opnode, above=0.35cm of unicycle-a, fill=white](bertram){\small\textsc{Bertram}};

\node[tnode,below=1.9cm of text-b](unicycle-copy){unicycle};

\draw [arrow] (riding-b) -- (riding-b |- bert-b.south);
\draw [arrow] (a-b) -- (a-b |- bert-b.south);
\draw [arrow] (un-b) -- (un-b |- bert-b.south);
\draw [arrow] (ic-b) -- (ic-b |- bert-b.south);
\draw [arrow] (y-b) -- (y-b |- bert-b.south);
\draw [arrow] (cle-b) -- (cle-b |- bert-b.south);
\draw [arrow] (is-b) -- (is-b |- bert-b.south);
\draw [arrow] (hard-b) -- (hard-b |- bert-b.south);

\draw [arrow] (riding-a) -- (riding-a |- bert-a.south);
\draw [arrow] (a-a) -- (a-a |- bert-a.south);
\draw [arrow] (is-a) -- (is-a |- bert-a.south);
\draw [arrow] (hard-a) -- (hard-a |- bert-a.south);
\draw [arrow] (bertram) -- (bertram-border.north);
\draw [arrow] (unicycle-a) -- (bertram);

\end{tikzpicture}
\caption{Top: Standard use of BERT. Bottom: Our proposal;
first \textsc{Bertram} learns an embedding for ``unicycle''
that replaces the WordPiece sequence. BERT is then run on
this improved input representation.}
\label{bertram-idea}
\end{figure} 

\section{Related Work}

Surface-form information (e.g., morphemes,
characters or character $n$-grams) is commonly used
to improve word representations. For
static word embeddings, this information can
either be injected into a given embedding space
\citep{luong2013better,pinter2017mimicking}, or a model can
directly be given access to it during training
\citep{bojanowski2016enriching,salle2018incorporating,edizel2019misspelling}. In
the area of contextualized representations, many
architectures employ subword segmentation methods
\citep[e.g.][]{radford2018improving,devlin2018bert,yang2019xlnet,liu2019roberta}.
Others use convolutional neural networks to directly
access character-level information
\citep{kim2016character,peters2018deep,baevski2019cloze}.

Complementary to surface form, another useful source of
information for understanding rare words are the contexts in
which they occur
\citep{lazaridou2017multimodal,herbelot2017high,khodak2018carte}. 
\citet{schick2019attentive,schick2018learning} show that combining
form and context leads to significantly better results than
using just one of the two.
While
all of these methods are 
bag-of-words models, \citet{liu-etal-2019-second} recently
proposed an architecture based
on  \emph{context2vec}  \citep{melamud-etal-2016-context2vec}. However,
in contrast to our work, they (i) do not incorporate
surface-form information and (ii) do not directly access the
hidden states of
context2vec, but instead simply use its output distribution. 

Several datasets focus on rare words, e.g., Stanford Rare
Word  \citep{luong2013better}, Definitional
Nonce  \citep{herbelot2017high}, and Contextual Rare
Word  \citep{khodak2018carte}. However,
unlike our rarified datasets, they are only suitable for evaluating \emph{uncontextualized} word representations.
Rarification is related to adversarial example generation \cite[e.g.][]{ebrahimi-etal-2018-hotflip}, which manipulates the input to change a model's prediction. We use a similar mechanism to determine which words in a given sentence are most important and replace them with rare synonyms.

\section{Model}

\subsection{Form-Context Model}
We first review
the basis for our new model,
the form-context model (FCM) \cite{schick2018learning}. Given a set of
$d$-dimensional high-quality embeddings for frequent words,
FCM induces embeddings for rare words that
are appropriate for the given embedding space. This is done
as follows: Given a word $w$ and a context $C$ in which it
occurs, a \emph{surface-form embedding}
$v_{(w,{C})}^\text{form} \in \mathbb{R}^d$ is obtained
by averaging over
embeddings of all character $n$-grams in $w$; the $n$-gram
embeddings are learned during training. Similarly, a
\emph{context embedding} $v_{(w,{C})}^\text{context} 
 \in \mathbb{R}^d$ is obtained by averaging over the embeddings
of all words in $C$. Finally, both embeddings
are combined using a gate 
\begin{align*}
g(v_{(w,{C})}^\text{form} , v_{(w,{C})}^\text{context} ) = \sigma(x^\top [v_{(w,{C})}^\text{form}  ; v_{(w,{C})}^\text{context}]  + y)
\end{align*}
with parameters $x \in \mathbb{R}^{2d}, y \in \mathbb{R}$
and $\sigma$ denoting the sigmoid function, allowing the
model to
decide
how to weight surface-form and context.
The final representation of $w$ is
then a weighted combination of form and context embeddings:
\begin{align*}
v_{(w,{C})} = \alpha \cdot (A v_{(w,{C})}^\text{context} + b) + (1 - \alpha) \cdot v_{(w,{C})}^\text{form}
\end{align*}
where $\alpha = g(v_{(w,C)}^\text{form}, v_{(w,C)}^\text{context})$ and $A \in \mathbb{R}^{d\times d}, b \in \mathbb{R}^d$ are parameters learned during training.

The context part of FCM is able to capture the broad
topic of rare words, but since it is a bag-of-words model,
it is not capable
of obtaining  a more concrete or detailed understanding
\citep[see][]{schick2018learning}.
%This is hardly surprising
%given the model's simplicity; it does, for example, make no
%use at all of the relative positions of context
%words.
Furthermore, the simple gating mechanism results in
only a shallow combination of form and context. That is, the
model is not able to combine form and context until the very
last step: While it can learn to weight form
and context components, the two embeddings (form and
context) do
not share any information and thus do not influence each
other.

\subsection{\textsc{Bertram}}

To overcome these limitations, we introduce \textsc{Bertram}, a model that
combines a pretrained BERT language model
\citep{devlin2018bert} with Attentive Mimicking
\citep{schick2019attentive}. We denote with
$e_{t}$ the (uncontextualized, i.e., first-layer) embedding
assigned to a (wordpiece) token
$t$ by BERT. Given a sequence of such uncontextualized
embeddings $\mathbf{e} = e_1, \ldots, e_n$, we denote by
$\textbf{h}_j(\textbf{e})$ the contextualized
representation of the $j$-th token at the final layer when the
model is given $\mathbf{e}$ as input.

Given a word $w$ and a context $C$ in which it occurs, let $\mathbf{t} = t_1, \ldots, t_{m}$ be the sequence obtained from $C$ by
(i) replacing $w$ with a \texttt{[MASK]} token and
(ii) tokenization (matching BERT's vocabulary);
 furthermore, let $i$ denote the index for which $t_i =
 \texttt{[MASK]}$.
 We experiment with three variants of \textsc{Bertram}: \textsc{Bertram-shallow}, \textsc{Bertram-replace} and \textsc{Bertram-add}.\footnote{We
   refer to these three \textsc{Bertram} configurations simply
   as \textsc{shallow}, \textsc{replace} and \textsc{add}.}

\paragraph{\textsc{shallow}.}
 Perhaps the simplest approach for obtaining a context embedding from $C$ using BERT is to define
\begin{align*}
v^\text{context}_{(w,C)} = \mathbf{h}_i(e_{t_1}, \ldots, e_{t_m})\,.
\end{align*}
This approach aligns well with BERT's pretraining objective of predicting likely substitutes for \texttt{[MASK]} tokens from their contexts.
The context
embedding $v^\text{context}_{(w,C)}$ is then combined with its form
counterpart as in FCM.

While this achieves our first goal of using a more
sophisticated context model that goes beyond bag-of-words,
it
still only combines form and
context in a shallow fashion.

%We thus refer to it as the
%\emph{shallow} variant of our model and investigate two
%alternative approaches (\emph{replace} and \emph{add}) that
%work as follows:

\paragraph{\textsc{replace}.}
  Before computing the context embedding, we replace the uncontextualized embedding of the \texttt{[MASK]} token with the word's surface-form embedding:
\[
v^\text{context}_{(w,C)} = \mathbf{h}_i(e_{t_1}, \mydots, e_{t_{i-1}}, v^\text{form}_{(w,C)} , e_{t_{i+1}}, \mydots, e_{t_m})\,.
\]
Our rationale for this is as follows:
During regular BERT pretraining, words chosen for prediction
are replaced with \texttt{[MASK]} tokens only 80\% of the
time and kept unchanged 10\% of the time. Thus, standard
pretrained BERT should
be able to make use of form embeddings presented this way as they provide a
strong signal with regards to how the ``correct'' embedding of $w$ may look like.

\paragraph{\textsc{add}.}
  Before computing the context embedding, we prepad the
  input with the surface-form embedding of $w$, followed by
  a colon ($e_:$):\footnote{
  We  experimented with  other prefixes, but
found that  this variant is best capable of
recovering $w$ at the masked position.}
\[
v^\text{context}_{(w,C)} = \mathbf{h}_{i+2}(v^\text{form}_{(w,C)} ,  e_:, e _{t_1}, \ldots, e_{t_m})\,.
\]
The intuition behind this third variant is that lexical definitions and explanations of a word $w$ are occasionally prefixed by ``$w$ :'' (e.g., in some online dictionaries). We assume that BERT has seen many definitional sentences of this kind during pretraining and is thus able to leverage surface-form information about $w$ presented this way.

\tikzset{
  tnode/.style={rectangle, inner sep=0.1cm, minimum height=3ex, text centered,text height=1.5ex, text depth=0.25ex},
  opnode/.style={draw, rectangle, rounded corners, minimum height=4ex, minimum width=4ex, text centered},
  arrow/.style={draw,->,>=stealth}
}

\begin{figure*}
\centering
\begin{tikzpicture}
% surface-form nodes
\node[tnode](Swas){\small $\langle$S$\rangle$was};
\node[tnode, right=0cm of Swas](wash){\small wash};
\node[tnode, right=0cm of wash](ngram-dots){\small $\ldots$};
\node[tnode, right=0cm of ngram-dots](lesS){\small les$\langle$S$\rangle$};
\node[fit=(Swas)(lesS), inner sep=0](ngrams){};
\node[tnode, above=1cm of ngrams](v-ngrams){$v^\text{form}_{(w, C_1)}$};

% context embedding nodes
\node[tnode,left=0.1cm of v-ngrams](v-cls){$e_\texttt{[CLS]}$};
\node[tnode,right=0.1cm of v-ngrams](v-colon){$e_\text{:}$};
\node[tnode,right=0.1cm of v-colon](v-other){$e_\text{other}$};
\node[tnode,right=0.1cm of v-other](v-mask){$e_\texttt{[MASK]}$};
\node[tnode,right=0.1cm of v-mask](v-such){$e_\text{such}$};
\node[tnode,right=0.1cm of v-such](v-as){$e_\text{as}$};
\node[tnode,right=0.1cm of v-as](v-trousers){$e_\text{trousers}$};
\node[tnode,right=0.1cm of v-trousers](v-context-dots){\small $\ldots$};

% context word nodes
\node[tnode, below=0.35 cm of v-colon](colon){\small :};
\node[tnode, below=0.35 cm of v-other](other){\small other};
\node[tnode, below=0.35 cm of v-mask](mask){\small \texttt{[MASK]}};
\node[tnode, below=0.35 cm of v-such](such){\small such};
\node[tnode, below=0.35 cm of v-as](as){\small as};
\node[tnode, below=0.35 cm of v-trousers](trousers){\small trousers};
\node[tnode, below=0.35 cm of v-context-dots](context-dots){\small $\ldots$};

\node[fit=(v-cls)(v-context-dots), inner sep=0](bert-input){};

\node[opnode, above=0.35cm of bert-input, minimum width=9.5cm, minimum height=0.8cm](bert){BERT};

%\node[tnode, above=1.7cm of v-mask](h-mask){$\mathbf{h}_5^{l_\text{max}}{(\textbf{e}_\text{add})}$};

%\node[fit=(h-mask), inner sep=0](bert-output){};

\node[opnode, above=1.5cm of v-mask](alpha){$A \cdot \dashedph + b$};
\node[tnode, above=0.3cm of alpha](final-output){$v_{(w,C_1)}$};

\node[draw, dashed, rounded corners, fit=(alpha)(bert)(Swas)](frame){};

\path [arrow] (Swas) -- (v-ngrams);
\path [arrow] (wash) -- (v-ngrams);
\path [arrow] (ngram-dots) -- (v-ngrams);
\path [arrow] (lesS) -- (v-ngrams);

\path [arrow] (colon) -- (v-colon);
\path [arrow] (mask) -- (v-mask);
\path [arrow] (such) -- (v-such);
\path [arrow] (as) -- (v-as);
\path [arrow] (trousers) -- (v-trousers);
\path [arrow] (other) -- (v-other);
\path [arrow] (context-dots) -- (v-context-dots);

\draw [arrow] (v-cls) -- (v-cls |- bert.south);
\draw [arrow] (v-ngrams) -- (v-ngrams |- bert.south);
\draw [arrow] (v-colon) -- (v-colon |- bert.south);
\draw [arrow] (v-mask) -- (v-mask |- bert.south);
\draw [arrow] (v-such) -- (v-such |- bert.south);
\draw [arrow] (v-as) -- (v-as |- bert.south);
\draw [arrow] (v-trousers) -- (v-trousers |- bert.south);
\draw [arrow] (v-other) -- (v-other |- bert.south);
\draw [arrow] (v-context-dots) -- (v-context-dots |- bert.south);

\draw [arrow] (alpha |- bert.north) -- (alpha);
%\path [arrow] (h-mask) -- (alpha);
\path [arrow] (alpha) -- (final-output);

\node[tnode, dashed, rounded corners, draw, minimum width=1.4cm, minimum height=1cm, right=1cm of frame](b1) {\small\textsc{Bertram}};
\node[tnode, dashed, rounded corners, minimum width=1.4cm, minimum height=1cm, right=0.2cm of b1](b2) {$\ldots$};
\node[tnode, dashed, rounded corners, draw, minimum width=1.4cm, minimum height=1cm, right=0.2cm of b2](b3) {\small\textsc{Bertram}};

\node[tnode, above=0.25cm of b1](vc1) {$v_{(w, C_1)}$};
\node[tnode, above=0.25cm of b2](vc2) {$\ldots$};
\node[tnode, above=0.25cm of b3](vc3) {$v_{(w, C_m)}$};

\node[tnode, below=1cm of b1](c1) {${(w, C_1)}$};
\node[tnode, below=1cm of b2](c2) {$\ldots$};
\node[tnode, below=1cm of b3](c3) {${(w, C_m)}$};

\node[fit=(vc1)(vc2)(vc3)(b1)(b2)(b3), inner sep=0](vs){};

\node[opnode, minimum width=4.5cm](am) at (alpha -| vs) {Attentive Mimicking};
\node[tnode, minimum width=4.5cm](final-output-g) at (final-output -| am) {$v_{(w, \mathcal{C})}$};

\draw [arrow] (c1) -- (b1);
\draw [arrow] (b1) -- (vc1);
\draw [arrow] (vc1) -- (am);

\draw [arrow] (c2) -- (b2);
\draw [arrow] (b2) -- (vc2);
\draw [arrow] (vc2) -- (am);

\draw [arrow] (c3) -- (b3);
\draw [arrow] (b3) -- (vc3);
\draw [arrow] (vc3) -- (am);

\draw [arrow] (am) -- (final-output-g);

\draw [dotted] (frame.north east) -- (b1.north west);
\draw [dotted] (frame.south east) -- (b1.south west);

\end{tikzpicture}
%\end{figure*} 

%\begin{figure}

%\includegraphics[width=\linewidth]{schematic}

\caption{
Schematic representation of \textsc{Bertram-add} processing the input word $w = \text{``washables''}$ given a single context $C_1=$ ``other washables such as trousers $\ldots$'' (left) and given multiple contexts $\mathcal{C} = \{C_1, \ldots, C_m\}$ (right)}
\label{bertram-architecture}

\end{figure*}
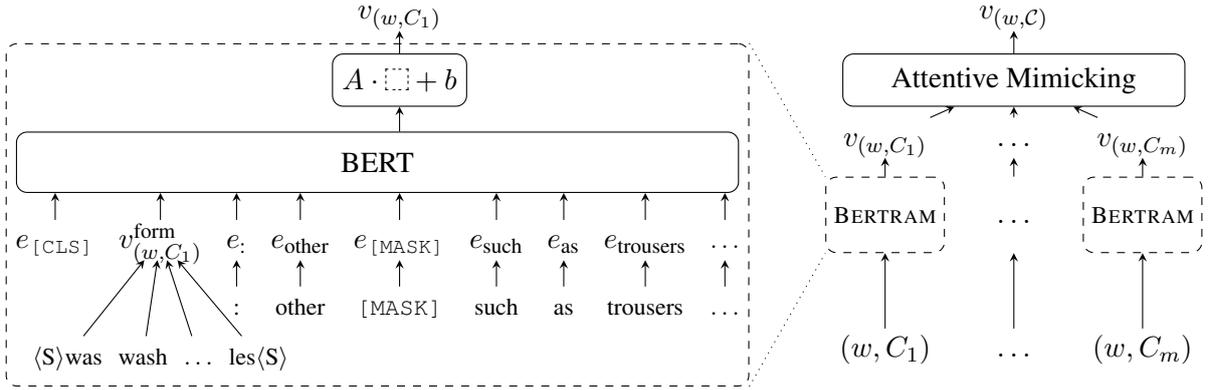

For both \textsc{replace} and \textsc{add}, surface-form information is directly and
deeply integrated into the computation of the context
embedding; thus, we do not require any gating
mechanism and directly set $v_{(w,C)} = A \cdot
v^\text{context}_{(w,C)} + b$. Figure~\ref{bertram-architecture} (left) shows how a single context is processed using \textsc{add}.

To exploit 
multiple contexts of a
word if available,
we follow the approach of
\citet{schick2019attentive} and
add an AM layer on top of our
model; see Figure~\ref{bertram-architecture}
(right). Given a set of contexts $\mathcal{C} =
\{C_1, \ldots, C_m\}$ and the corresponding embeddings
$v_{(w,C_1)}, \ldots, v_{(w,C_m)}$, AM applies a
self-attention mechanism to all embeddings, allowing the
model to distinguish informative from uninformative contexts.
The final embedding $v_{(w, \mathcal{C})}$ is then a
weighted combination of all embeddings:
\[
v_{(w, \mathcal{C})} = \sum\nolimits_{i=1}^m \rho_i \cdot v_{(w, C_i)}
\]
where the
self-attention layer determines
the weights $\rho_i$ subject to $\sum_{i=1}^m \rho_i = 1$.
For further details, see
\citet{schick2019attentive}.

\subsection{Training}
\label{section-bertram-training}

Like previous work, we use \emph{mimicking} \citep{pinter2017mimicking} as a training objective. That is, given a frequent word $w$ with known embedding $e_w$ and a set of corresponding contexts $\mathcal{C}$, \textsc{Bertram} is trained to minimize $\|e_w - v_{(w, \mathcal{C})}\|^2$.

Training \textsc{Bertram} end-to-end
is costly:
the cost of processing a single training instance
$(w,\mathcal{C})$ with $\mathcal{C} = \{C_1, \ldots, C_m\}$ is the same as processing an entire batch
of $m$ examples in standard BERT.
Therefore,  we resort to the following three-stage
training process:
\begin{enumerate}
\item We train only the context part, minimizing $\| e_w - A
  \cdot (\sum_{i=1}^m \rho_i \cdot v^\text{context}_{(w, {C_i})}) + b \|^2$ where $\rho_i$ is the weight assigned to each context $C_i$ through the AM layer. Regardless of the selected \textsc{Bertram} variant, the
  context embedding is always obtained using \textsc{shallow} in this stage. Furthermore, only $A$, $b$ and all parameters of the AM layer are optimized.
\item We train only the form part (i.e., only the $n$-gram embeddings); our loss for a
    single example $(w, \mathcal{C})$ is $\| e_w -
    v^\text{form}_{(w, \mathcal{C})} \|^2$. Training in this stage is completely detached from the underlying BERT model.
\item In the third stage, we combine the pretrained form-only and context-only
  models and train all parameters. The first two stages
are only run once and then used for all three \textsc{Bertram} variants
%are agnostic of the chosen
because context and form are
trained in isolation.
 The third stage must be run for each variant separately.
\end{enumerate}
We freeze all of BERT's parameters during training as we -- somewhat surprisingly -- found that this
slightly improves the model's performance while
speeding up training. For \textsc{add}, we additionally found it helpful to freeze the form part in the third training stage.
Importantly, for
the first two stages of our training procedure, we do not
have to backpropagate through BERT to
obtain all required gradients, drastically increasing the
training speed.

\section{Dataset Rarification}
\label{sec:section-rwgen}

The ideal dataset for
measuring the quality of rare word representations would be
one for which
the accuracy of
a model with no understanding of rare words is
0\% whereas the accuracy of a model that perfectly understands rare words is 100\%.
Unfortunately,
existing datasets do not satisfy this desideratum, 
not least because
rare words -- by their nature -- occur rarely.

This does not mean that rare words are not important: As we
shift our focus in NLP from words and sentences as the main
unit of processing to larger units like paragraphs and
documents, rare words will occur in a high proportion of
such larger ``evaluation units''. Rare words are also clearly a
hallmark of human language competence, which should be the
ultimate goal of NLP. Our work is part of a trend that
sees a need for evaluation tasks in NLP that are more
ambitious than what we have now.\footnote{Cf.\ 
\citep{bowman19medium}: ``If we want to be able to establish fair benchmarks that
encourage future progress toward robust, human-like language
understanding, we'll need to get
better at creating clean, challenging, and realistic test
datasets.''}

%; even when they do, they often of negligible importance.

To create more challenging datasets,
we use \emph{rarification}, a procedure that automatically transforms existing text classification
datasets in such a way that rare words become important.
We require a pretrained language model $M$ as a
baseline, an arbitrary text classification dataset
$\mathcal{D}$ containing labeled instances $(\mathbf{x},
y)$ and a \emph{substitution dictionary} $S$, mapping each
word $w$ to a set of rare synonyms $S(w)$.  Given these
ingredients, our procedure consists of three steps: (i)
splitting the dataset into a train set and a set of test
candidates, (ii) training the baseline model on the train
set and (iii) modifying a subset of the test candidates to
generate the final test set.

\paragraph{Dataset Splitting.}
We partition $\mathcal{D}$ into a training set
$\mathcal{D}_\text{train}$ and a set of \emph{test
candidates},
$\mathcal{D}_\text{cand}$. $\mathcal{D}_\text{cand}$
contains all instances $(\mathbf{x},y) \in \mathcal{D}$ such
that for at least one word $w$ in $\mathbf{x}$,
$S(w) \neq \emptyset$ -- subject to the constraint that the training set contains at least one third of the entire data.

\paragraph{Baseline Training.} We finetune $M$ on $\mathcal{D}_\text{train}$. Let $(\mathbf{x}, y) \in \mathcal{D}_\text{train}$ where $\mathbf{x} = w_1, \ldots, w_n$ is a sequence of words. We deviate from the finetuning procedure of \citet{devlin2018bert} in three respects:

\begin{itemize}
\item We randomly replace 5\% of all words in $\mathbf{x}$ with a \texttt{[MASK]} token. This allows the model to cope with missing or unknown words, a prerequisite for our final test set generation.
\item As an alternative to overwriting the language model's
uncontextualized embeddings for rare words, we also want to
allow models to \emph{add} an alternative representation
during test time, in which case we simply separate both
representations by a slash (cf.\ \secref{section-downstream-tasks}).
To accustom the language model to this duplication of words, we replace each word $w_i$ with ``$w_i$ / $w_i$'' with a probability of 10\%. To make sure that the model does not simply learn to always focus on the first instance during training, we randomly mask each of the two repetitions with probability 25\%.
\item We do not finetune the model's embedding layer.
We found that this does not hurt performance, an observation in line with recent findings of \citet{lee2019would}.
\end{itemize}

\paragraph{Test Set Generation.}
Let $p(y \mid \mathbf{x})$ be the probability that the
finetuned model $M$ assigns to class $y$ given input
$\mathbf{x}$, and
$
M(\mathbf{x}) = \argmax_{y \in \mathcal{Y}} p(y \mid \mathbf{x})
$ be the model's prediction for input $\mathbf{x}$ where $\mathcal{Y}$ denotes the set of all labels. 
For generating our test set, we only consider candidates that are classified \emph{correctly} by the baseline model, i.e., candidates $(\mathbf{x}, y) \in \mathcal{D}_\text{cand}$ with $M(\mathbf{x}) = y$. For each such entry, let $\mathbf{x} = w_1, \ldots, w_n$ and let $\mathbf{x}_{w_i = t}$ be the sequence obtained from $\mathbf{x}$ by replacing $w_i$ with $t$. We compute
\[
{w}_i =\argmin_{w_j: S(w_j) \neq \emptyset} p(y \mid \mathbf{x}_{w_j = \texttt{[MASK]}}),
\]
i.e., we select the word $w_i$ whose masking pushes the
model's prediction the farthest away from the correct
label. If removing this word already changes the model's
prediction -- that is, $M(\mathbf{x}_{w_i =
  \texttt{[MASK]}}) \neq y$ --, we select a random rare
synonym $\hat{w}_i \in S(w_i)$ and add $(\mathbf{x}_{w_i =
  \hat{w}_i}, y)$ to the test set. Otherwise, we repeat the
above procedure; if the label still has not changed after
masking up to $5$ words, we discard the candidate.
Each instance $(\mathbf{x}_{w_{i_1} = \hat{w}_{i_1}, \ldots,
  w_{i_k} = \hat{w}_{i_k} }, y)$ of the resulting test set has the following properties:
\begin{itemize}
\item If each $w_{i_j}$ is replaced by \texttt{[MASK]}, the entry is classified incorrectly by $M$. In other words, understanding the words $w_{i_j}$ is necessary for $M$ to determine the correct label.

\item If the model's internal representation of each
$\hat{w}_{i_j}$ is sufficiently similar to its
representation of $w_{i_j}$, the entry is classified
correctly by $M$. That is, if the model is able to
understand the rare words $\hat{w}_{i_j}$ and to identify
them as synonyms of ${w_{i_j}}$, it will predict the correct label.
\end{itemize}

Note that the test set is closely coupled to the baseline
model $M$ because we select the words to be replaced based
on $M$'s predictions. Importantly, however, the model
is never queried with any rare synonym during test set
generation, so its representations of rare words
are \emph{not} taken into account for creating the test
set. Thus, while the test set is not suitable for comparing
$M$ with an entirely different model $M'$, it allows us to
compare various strategies for representing rare words in
the embedding space of $M$. 
Definitional Nonce
\citep{herbelot2017high}
is subject to a similar constraint:
it is tied to a specific
(uncontextualized) embedding space based on Word2Vec \citep{Mikolov2013}.

\section{Evaluation}

\subsection{Setup}

For our evaluation of \textsc{Bertram}, we follow the experimental setup of \citet{schick2019ota}. We experiment with integrating \textsc{Bertram} both into BERT$_\text{base}$ and RoBERTa$_\text{large}$ \citep{liu2019roberta}. Throughout our experiments, when \textsc{Bertram} is used to provide input representations for one of the two models, we use the same model as \textsc{Bertram}'s underlying language model. Further training specifications can be found in Appendix~\ref{appendix-training-parameters}.

\begin{table}
\small
\begin{tabularx}{\linewidth}{Xcc}
%\begin{tabular}{lcc}
\toprule
%& \multicolumn{2}{c}{\textbf{MRR}} \\
%\cmidrule(lr){2-3}
\textbf{Model} & \textsc{rare} & \textsc{medium} \\
\midrule
BERT (base) & 0.112 & 0.234 \\
\ + AM \citep{schick2019ota} & 0.251 & 0.267 \\
\ + \textsc{Bertram-shallow} & 0.250 & 0.246 \\
\ + \textsc{Bertram-replace} & 0.155 & 0.216 \\
\ + \textsc{Bertram-add} & \underline{\textbf{0.269}} & \underline{\textbf{0.367}} \\
BERT (large) & 0.143 & 0.264 \\
\cdashlinelr{1-3}
RoBERTa (large) & 0.270 & 0.275 \\
\ + \textsc{Bertram-add} & \underline{\textbf{0.306}} & \underline{\textbf{0.323}} \\
\bottomrule
\end{tabularx}
%\end{tabular}
\caption{MRR on WNLaMPro test for baseline models and various \textsc{Bertram} configurations. Best results per base model are underlined, results that do not differ significantly from the best results in a paired t-test ($p < 0.05$) are bold.}
\label{results-wnlampro}
\end{table}

While BERT was trained on BookCorpus \cite{zhu2015aligning}
and a large Wikipedia dump, we follow previous work and
train \textsc{Bertram} only on the much smaller Westbury
Wikipedia Corpus (WWC) \citep{shaoul2010westbury}; this of
course gives BERT a clear advantage over \textsc{Bertram}. This advantage is even more pronounced when comparing \textsc{Bertram} with RoBERTa, which is trained on a corpus that is an order of magnitude larger than the original BERT corpus.
We try to at least partially compensate for this as follows:
In our downstream
task experiments, we gather the set of contexts $\mathcal{C}$
for each word from WWC+BookCorpus
during inference.\footnote{We recreate BookCorpus with the script at
  \url{github.com/soskek/bookcorpus}.
We refer to the
joined corpus of WWC and BookCorpus as WWC+BookCorpus.} 

\subsection{WNLaMPro}

We evaluate \textsc{Bertram} on the WNLaMPro dataset \citep{schick2019ota}. This dataset consists of cloze-style phrases like 
``A \emph{lingonberry} is a \_\_\_\_.''
and the task is to correctly fill the slot (\_\_\_\_) with
one of several acceptable target words (e.g., ``fruit'',
``bush'' or ``berry''), which requires understanding of the
meaning of the phrase's \emph{keyword} (``lingonberry'' in
the example). As the goal of this dataset is to probe a
language model's ability to understand rare words
\emph{without} any task-specific finetuning,
\citet{schick2019ota} do not provide a training set.  The
dataset is partitioned into three subsets
based on the keyword's frequency in WWC:
\textsc{rare}
(occurring fewer than 10 times)
\textsc{medium}
(occurring between 10 and 100 times),
and \textsc{frequent}
(all remaining words).

\begin{table}
\small
\begin{tabularx}{\linewidth}{lX}
\toprule
{\textbf{Task}} & {\textbf{Entry}} \\
\midrule
\footnotesize{MNLI} & i think i will go finish up my \sout{laundry} \textbf{washables}.\\

\footnotesize{AG's} & [\ldots] stake will \sout{improve} \textbf{meliorate} symantec's consulting contacts [\ldots]\\

\footnotesize{DBPedia} & yukijiro hotaru [\ldots] is a \sout{japanese} \textbf{nipponese} \sout{actor} \textbf{histrion}.\\

\footnotesize{MNLI} & a smart person is \sout{often} \textbf{ofttimes} correct in their \sout{answers} \textbf{ansers}.\\

\footnotesize{MNLI} & the southwest has a lot of farming and \sout{vineyards} \textbf{vineries} that make \sout{excellent} \textbf{fantabulous} merlot.\\

\bottomrule
\end{tabularx}
\caption{Examples from rarified datasets.
Crossed out: replaced words. Bold: replacements.}
\label{substitution-dict}
\end{table}

\begin{table*}
\small
\begin{tabularx}{\linewidth}{Xccccccccccc}
%\begin{tabular}{lccccccccc}
\toprule
& \multicolumn{3}{c}{\textbf{MNLI}} &
& \multicolumn{3}{c}{\textbf{AG's News}} & 
& \multicolumn{3}{c}{\textbf{DBPedia}} \\
\cmidrule(lr){2-4}
\cmidrule(lr){6-8}
\cmidrule(lr){10-12}
\textbf{Model} & All & Msp & WN && All & Msp & WN  && All & Msp & WN \\
\midrule
BERT (base)
& 50.5 & 49.1 & 53.4 && 56.5 & 54.8 & 61.9 && 49.3 & 46.0 & 57.6 \\

\ + Mimick \citep{pinter2017mimicking}
& 37.2 & 38.2 & 38.7 && 45.3 & 43.9 & 50.5 && 36.5 & 35.8 & 41.1 \\

\ + A La Carte \citep{khodak2018carte}
& 44.6 & 45.7 & 46.1  && 52.4 & 53.7 & 56.1 && 51.1 & 48.7 & 59.3 \\

\ + AM \citep{schick2019ota}				
& 50.9 & 50.7 & 53.6 && 58.9 & 59.8 & 62.6 && 60.7 & 63.1 & 62.8 \\

\ + \textsc{Bertram} 				
& 53.3 & 52.5 & 55.6  && \textbf{62.1} & \textbf{63.1} & \textbf{65.3} && 64.2 & 67.9 & 64.1  \\

\ + \textsc{Bertram-slash}			
& 56.4 & 55.3 & 58.6 && \underline{\textbf{62.9}} & \underline{\textbf{63.3}} & \textbf{65.3} && 65.7 & 67.3 & 67.2 \\

\ + \textsc{Bertram-slash} + \textsc{indomain}  
 & \underline{\textbf{59.8}} & \underline{\textbf{57.3}} & \underline{\textbf{62.7}} && \textbf{62.5} & \textbf{62.1} & \underline{\textbf{66.6}} && \underline{\textbf{74.2}} & \underline{\textbf{74.8}} & \underline{\textbf{76.7}} \\

\cdashlinelr{1-12}

RoBERTa (large)
& 67.3 & 68.7 & 68.4 && 63.7 & 68.1 & 65.7 && 65.5 & 67.3 & 66.6 \\
\ + \textsc{Bertram-slash}
& 70.1 & \textbf{71.5} & 70.9 && 64.6 & 68.4 & 64.9 && 71.9 & 73.8 & 73.9 \\
\ + \textsc{Bertram-slash} + \textsc{indomain}
& \underline{\textbf{71.7}} & \underline{\textbf{71.9}} & \underline{\textbf{73.2}} && \underline{\textbf{68.1}} & \underline{\textbf{71.9}} & \underline{\textbf{69.0}} && \underline{\textbf{76.0}} & \underline{\textbf{78.8}} & \underline{\textbf{77.3}} \\

\bottomrule
\end{tabularx}
%\end{tabular}
\caption{Accuracy of standalone BERT and RoBERTa,
  various baselines and \textsc{Bertram} on
  rarified MNLI, AG's News and DBPedia.
The five \textsc{Bertram} instances are \textsc{Bertram-add}.
  Best results per baseline model are underlined, results that do not differ significantly from 
  the best results in a two-sided binomial test ($p < 0.05$) are bold.
Msp/WN: subset of instances containing at least one
  misspelling/synonym.
All: all instances.}
\label{results-downstream}
\end{table*} 

For our evaluation, we compare the performance of a
standalone BERT (or RoBERTa) model with one that uses
\textsc{Bertram} as shown in Figure~\ref{bertram-idea} (bottom).
%; in the following, we refer to this combination as BERT+\textsc{Bertram} or, in short, $\textsc{Bert}2\textsc{ram}$.
As our focus is to improve representations for rare words,
we evaluate our model only on WNLaMPro
\textsc{rare} and
\textsc{medium}. Table~\ref{results-wnlampro} gives results;
our
measure is mean reciprocal rank (MRR).  We see that
supplementing BERT with any of the proposed
methods results in noticeable improvements for the
\textsc{rare} subset, with  \textsc{add} clearly outperforming
\textsc{shallow} and \textsc{replace}.  Moreover,  \textsc{add}
performs surprisingly well for more frequent
words, improving the score for WNLaMPro-\textsc{medium} by
58\% compared to BERT$_\text{base}$ and 37\% compared to
Attentive Mimicking. This makes sense considering that
the key enhancement of
\textsc{Bertram} over AM lies in improving context representations
and interconnection of form and context; the more
contexts are given, the more this comes into
play. Noticeably, despite being both based on and integrated
into a BERT$_\text{base}$ model, our architecture even
outperforms BERT$_\text{large}$  by a
large margin. While RoBERTa performs much better than BERT
on WNLaMPro, \textsc{Bertram} still significantly improves
results for both rare and medium frequency words. As it
performs best for both the \textsc{rare} and \textsc{medium}
subset,  we always use the \textsc{add} configuration of
\textsc{Bertram} in the following experiments.

\iffalse
\begin{table*}
\small
\begin{tabularx}{\linewidth}{Xl}
\toprule
\textbf{Entry} & \textbf{Replacements} \\
\midrule
i think i will go finish up my \textbf{washables} . & laundry $\mapsto$ washables \\[0.1cm]

[\ldots] stake will \textbf{meliorate} symantec ' s consulting contacts [\ldots] & improve $\mapsto$ meliorate \\[0.1cm]

yukijiro hotaru yukijiro hotaru [\ldots] is a \textbf{nipponese histrion} . & japanese $\mapsto$ nipponese, actor $\mapsto$ histrion \\[0.1cm]

a smart person is \textbf{ofttimes} correct in their \textbf{ansers} . & often $\mapsto$ ofttimes, answers $\mapsto$ ansers \\[0.1cm]

the southwest has a lot of farming and \textbf{vineries} that make \textbf{fantabulous} merlot . & vineyards $\mapsto$ vineries, excellent $\mapsto$ fantabulous \\

\bottomrule
\end{tabularx}
\caption{Example entries from the datasets obtained through our procedure}
\label{substitution-dict-alt}
\end{table*}
\fi

% Examples of so-obtained replacements include ``unseeable'', ``lawlessly'' and ``scarceness'' for ``invisible'', ``illegally'' and ``scarcity'', respectively. 

\subsection{Downstream Task Datasets}
\label{sec:section-downstream-tasks}

To measure the effect of adding \textsc{Bertram} to a pretrained deep language model on
downstream tasks, we
rarify
(cf.\ \secref{section-rwgen}) the following three datasets:
\begin{itemize}
\item MNLI \citep{williams2018mnli}, a natural language inference dataset where given two sentences $a$ and $b$, the task is to decide whether $a$ entails $b$, $a$ and $b$ contradict each other or neither;
\item AG's News \citep{zhang2015character}, a news classification dataset with four different categories (\emph{world}, \emph{sports}, \emph{business} and \emph{science/tech});
\item DBPedia \citep{lehmann2015dbpedia}, an ontology dataset with 14 classes (e.g., \emph{company}, \emph{artist}) that have to be identified from text snippets.
\end{itemize}

For all three datasets,
we create rarified instances both using BERT$_\text{base}$ and RoBERTa$_\text{large}$ as a baseline model and build the
substitution dictionary $S$ using the synonym relation of
WordNet \citep{miller1995wordnet} and the \emph{pattern}
library \citep{smedt2012pattern} to make sure that all
synonyms have consistent parts of speech. Furthermore, we only consider synonyms for each word's most frequent sense; this filters out much noise and improves the quality of the created sentences.
In addition to WordNet, we use the misspelling
dataset of \citet{edizel2019misspelling}. 
To prevent misspellings from
dominating the resulting datasets, we only assign
misspelling-based substitutes to randomly selected 10\% of
the words contained in each sentence.  Motivated by the
results on WNLaMPro-\textsc{medium}, we consider every word
that occurs less than 100 times in WWC+BookCorpus as being
rare.
Example entries from the rarified datasets obtained using BERT$_\text{base}$ as a baseline model can be seen
in Table~\ref{substitution-dict}. The average number of words replaced with synonyms or misspellings is $1.38$, $1.82$ and $2.34$ for MNLI, AG's News and DBPedia, respectively. 

Our default way of
injecting \textsc{Bertram} embeddings into the baseline
model is to replace the sequence of uncontextualized
subword token embeddings for a given rare word with
its \textsc{Bertram}-based embedding
(Figure~\ref{bertram-idea}, bottom). That is, given a
sequence of uncontextualized token embeddings $\mathbf{e} =
e_1, \ldots, e_n$ where $e_{i}, \ldots, e_{j}$ with
$1 \leq i \leq j \leq n$ is the sequence of
embeddings for a single rare word $w$ with \textsc{Bertram}-based embedding $v_{(w,\mathcal{C})}$, we replace
$\mathbf{e}$ with
\[
 \mathbf{e}' = e_1, \ldots, e_{i-1}, v_{(w,\mathcal{C})}, e_{j+1}, \ldots, e_n\,.
 \]
 
As an alternative to replacing the original sequence of subword embeddings for a given rare word, we also consider \textsc{Bertram-slash}, a configuration where
the \textsc{Bertram}-based embedding is simply added and both
representations are separated using a single slash:
\[
  \mathbf{e}_\textsc{slash} = e_1, \ldots, e_{j},
  e_{\text{/}}, v_{(w,\mathcal{C})}, e_{j+1}, \ldots, e_n\,. %\label{slash}
\]
The intuition behind this variant is that in BERT's pretraining corpus, a slash is often used to separate two variants of the same word (e.g., ``useable / usable'') or two closely related concepts (e.g., ``company / organization'', ``web-based / cloud'') and thus, BERT should be able to understand that both $e_i,\ldots,e_j$ and $v_{(w, \mathcal{C})}$ refer to the same entity. We therefore surmise that whenever some information is encoded in one representation but not in the other, giving BERT both representations is helpful.

By default, the set of contexts $\mathcal{C}$ for each word
 is obtained by collecting all sentences
 from WWC+BookCorpus in which it occurs. We also try a
 variant where we add in-domain contexts by giving \textsc{Bertram}
 access to all texts (but not labels) found in the test set; we refer to this
 variant as \textsc{indomain}.\footnote{For the MNLI
   dataset, which consists of text pairs $(a,b)$, we treat
   $a$ and $b$ as separate contexts.} Our motivation for including this
variant is as follows: Moving from the training stage of a model to its production
use often causes a slight domain shift. This is turn leads to an increased number
of input sentences containing words that did not -- or only very rarely -- appear in the training data.
However, such input sentences can easily be collected as additional unlabeled examples during production use. While
there is no straightforward way to leverage these unlabeled examples with an already
finetuned BERT model, \textsc{Bertram} can easily make use of them without
requiring any labels or any further training: They can simply be included as additional contexts during inference. As this gives \textsc{Bertram}
a slight advantage, we also report results for all configurations without using
indomain data. Importantly, adding indomain data increases the number of contexts for 
more than 90\% of all rare words by at most 3, meaning that they can still be considered rare despite the additional indomain contexts.
   
 Table~\ref{results-downstream} reports, for each task, the
 accuracy on the entire dataset (All) as well as scores
 obtained considering only instances where at least one word
 was replaced by a misspelling (Msp) or a WordNet synonym
 (WN), respectively.\footnote{Note that results for BERT and
   RoBERTa are only loosely comparable because the datasets
   generated from both baseline models through rarification
   are different.} Consistent with results on WNLaMPro,
 combining BERT with \textsc{Bertram} consistently outperforms both a
 standalone BERT model and one combined with various
 baseline models. Using the \textsc{slash}
 variant brings improvements across all datasets
 as does 
 adding \textsc{indomain} contexts
(exception: BERT/AG's News).
This makes sense considering that for a
 rare word, every single additional context can be crucial
 for gaining a deeper understanding. Correspondingly, it is
 not surprising that the benefit of adding
 \textsc{Bertram} to RoBERTa is less pronounced, because
 \textsc{Bertram} uses only a fraction of the contexts
 available to RoBERTa during pretraining. Nonetheless,
 adding \textsc{Bertram} significantly improves RoBERTa's
 accuracy for all three datasets both with and without adding 
 \textsc{indomain} contexts.

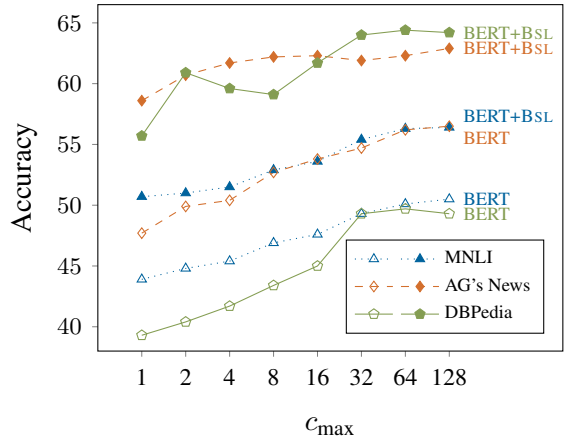
\begin{figure}
\centering
\pgfplotsset{
    compat=newest,
    /pgfplots/legend image code/.code={%
        \draw[mark repeat=2,mark phase=2,#1] 
            plot coordinates {
                (0cm,0cm) 
                (0.25cm,0cm)
                (0.5cm,0cm)
            };
    },
}
\begin{tikzpicture}
\begin{axis}[
	cycle list name=color list,
	xlabel={$c_\text{max}$\vphantom{Word Counts}},
		ylabel={Accuracy},
    ymin = 38,
    ymax = 66.5,
    xmin = 0.5,
    xmax = 768,
    xmode = log,
    xtick pos=left,
    ytick pos=left,
    log basis x={2},
        ylabel near ticks,
        xlabel near ticks,
    xtick = data,
    xticklabels = {1$\vphantom{[)}$,2$\vphantom{[)}$,4$\vphantom{[)}$,8$\vphantom{[)}$,16$\vphantom{[)}$,32$\vphantom{[)}$,64$\vphantom{[)}$,128$\vphantom{[)}$},
    ytick = {40, 45, 50, 55, 60, 65},
    tick align=outside,
          major tick length=0.075cm,
    width = \linewidth,
    height = 0.25\textheight,
    log ticks with fixed point,
    x tick label style={/pgf/number format/1000 sep=\,},
    legend style={at={(0.96,0.05)},anchor=south east, font=\scriptsize},
    legend cell align=left,
    legend columns=2,
    legend entries={$\!$,MNLI,$\!$,AG's News,$\!$,DBPedia},
    tick label style={font=\footnotesize}
]
\addlegendimage{mark=triangle,mark options={solid},dotted,c0}
\addlegendimage{mark=triangle*,mark options={solid},dotted,c0}
\addlegendimage{mark=diamond,mark options={solid},dashed,c1}
\addlegendimage{mark=diamond*,mark options={solid},dashed,c1}
\addlegendimage{mark=pentagon,c4}
\addlegendimage{mark=pentagon*,c4} % alternative: no markers

\addplot[mark=triangle, dotted, c0, mark options={solid}] coordinates {
(1,43.9)
(2,44.8)
(4,45.4)
(8,46.9)
(16,47.6)
(32,49.3)
(64,50.1)
(128,50.5) 
} node[right,pos=1,xshift=0.05cm]{\scriptsize{BERT}};

\addplot[mark=triangle*, dotted, c0, mark options={solid}] coordinates {
(1,50.7)
(2,51.0)
(4,51.5)
(8,52.9)
(16,53.6)
(32,55.4)
(64,56.3)
(128,56.4)
} node[right,pos=1,xshift=0.05cm,yshift=0.15cm]{\scriptsize{BERT+\textsc{Bsl}}};

\addplot[mark=diamond, dashed, c1, mark options={solid}] coordinates {
(1,47.7)
(2,49.9)
(4,50.4)
(8,52.7)
(16,53.8)
(32,54.7)
(64,56.2)
(128,56.5) 
} node[right,pos=1,xshift=0.05cm,yshift=-0.15cm]{\scriptsize{BERT}};

\addplot+[mark=diamond*, dashed, c1, mark options={solid}] coordinates {
(1,58.6)
(2,60.7)
(4,61.7)
(8,62.2)
(16,62.3)
(32,61.9)
(64,62.3)
(128,62.9)
} node[right,pos=1,xshift=0.05cm]{\scriptsize{BERT+\textsc{Bsl}}};

\addplot[mark=pentagon, c4, mark options={solid}] coordinates {
(1,39.3)
(2,40.4)
(4,41.7)
(8,43.4)
(16,45.0)
(32,49.3)
(64,49.7)
(128,49.3)
} node[right,pos=1,xshift=0.05cm]{\scriptsize{BERT}};

\addplot+[mark=pentagon*, c4, mark options={solid}] coordinates {
(1,55.7)
(2,60.9)
(4,59.6)
(8,59.1)
(16,61.7)
(32,64.0)
(64,64.4)
(128,64.2)
} node[right,pos=1,xshift=0.05cm]{\scriptsize{BERT+\textsc{Bsl}}};

\end{axis}
\end{tikzpicture}
\caption{BERT vs. BERT combined with \textsc{Bertram-slash} (BERT+\textsc{Bsl}) on three downstream tasks for varying maximum numbers of contexts $c_\text{max}$}
\label{results-freq}
\end{figure}

To further understand for which words using \textsc{Bertram}
is helpful, Figure~\ref{results-freq} looks at the
accuracy of BERT$_\text{base}$ both with and without \textsc{Bertram}  as a function of word frequency. That is, we
compute the accuracy scores for both models when considering
only entries $(\mathbf{x}_{w_{i_1} = \hat{w}_{i_1}, \ldots,
  w_{i_k} = \hat{w}_{i_k} }, y)$ where each substituted word
$\hat{w}_{i_j}$ occurs less than $c_\text{max}$ times in
WWC+BookCorpus, for different values of $c_\text{max}$. As one
would expect, $c_\text{max}$ is positively correlated with
the accuracies of both models, showing that the rarer a word
is, the harder it is to understand. 
Interestingly, the gap between
standalone BERT and BERT with \textsc{Bertram} remains more or less constant
regardless of $c_\text{max}$. This suggests that
using \textsc{Bertram} may even be helpful for more frequent
words.

To investigate this hypothesis, we perform another rarification of
MNLI that differs from the previous rarification
in two respects. First, we increase the threshold
for a word to count as rare from 100 to 1000. Second, as
this means that we have more WordNet synonyms available, we
do not use the misspelling dictionary
\cite{edizel2019misspelling} for substitution. We refer to the
resulting datasets for BERT$_\text{base}$ and RoBERTa$_\text{large}$ as \emph{MNLI-1000}.

\begin{figure}
\centering
\begin{tikzpicture}
\begin{axis}[
	cycle list name=color list,
	xlabel={Word counts\vphantom{$c_\text{max}$}},
		ylabel={Accuracy improvement},
    ymin = -2,
    ymax = 15.5,
    xmin = 0.72,
    xmax = 4.28,
    ytick = {0, 2, 4, 6, 8, 10},
    xtick = {1, 2, 3, 4},
    xticklabels = {$[0{,}125)$, $[125{,}250)$, $[250{,}500)$, $[500{,}\infty)$},
    xtick pos=left,
    ytick pos=left,
    ylabel near ticks,
    xlabel near ticks,
    tick align=outside,
    major tick length=0.075cm,
    width = \linewidth,
    height = 0.25\textheight,
    x tick label style={/pgf/number format/1000 sep=},
    legend style={at={(0.5,0.95)},anchor=north, font=\scriptsize},
    legend cell align=left,
    legend columns=2,
    tick label style={font=\footnotesize}
]

\addplot[mark=square, c0, mark options={solid}] coordinates {
(1,6.9)
(2,6.0)
(3,9.6)
(4,6.6)
};
\addlegendentry{BERT+\textsc{Bsl}}

\addplot[mark=o, c1, mark options={solid}] coordinates {
(1,2.2)
(2,5.6)
(3,1.3)
(4,-0.6)
};
\addlegendentry{RoBERTa+\textsc{Bsl}}

\addplot[dashed, mark=square*, c0, mark options={solid}] coordinates {
(1,9.9)
(2,9.3)
(3,4.5)
(4,6.4)
};
\addlegendentry{BERT+\textsc{Bsl}+ID}

\addplot[dashed, mark=*, c1, mark options={solid}] coordinates {
(1,4.8)
(2,6.6)
(3,1.3)
(4,0.4)
};
\addlegendentry{RoBERTa+\textsc{Bsl}+ID}

\addplot[mark=none, dotted, black, mark options={solid}] coordinates {
(0.5,0)
(4.5,0)
};
\end{axis}
\end{tikzpicture}
\caption{Improvements for BERT (base) and RoBERTa (large) when adding \textsc{Bertram-slash} (+\textsc{Bsl}) or \textsc{Bertram-slash + indomain} (+\textsc{Bsl}+ID) on MNLI-1000}
\label{results-mnli-1000}
\end{figure}
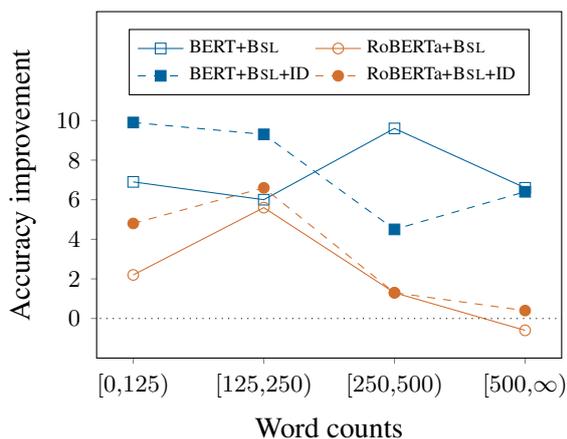

Figure~\ref{results-mnli-1000} shows
results on MNLI-1000
for various rare word
frequency ranges.  For each value $[c_0, c_1)$ on the $x$-axis, the $y$-axis shows improvement in accuracy compared to standalone BERT or RoBERTa when only dataset entries are considered for which each rarified word occurs between $c_0$ (inclusively) and $c_1$ (exclusively) times in WWC+BooksCorpus. We see that for
words with frequency less than 125, the improvement in accuracy
remains similar even without using misspellings as
another source of substitutions. Interestingly, for every
single interval of rare word counts
considered, adding \textsc{Bertram-slash}
to BERT considerably improves its accuracy. For RoBERTa, adding \textsc{Bertram} brings improvements only for words occurring less than 500 times.
While using \textsc{indomain} data is
beneficial for rare words -- simply because it gives us
additional contexts for these words --, when considering
only words that occur at least 250 times in
WWC+BookCorpus, adding \textsc{indomain}
contexts does not help.

\section{Conclusion}

We have introduced \textsc{Bertram}, a novel architecture
for inducing high-quality representations for rare words
in BERT's and RoBERTa's embedding spaces. This is achieved by employing a
powerful pretrained language model and deeply integrating
surface-form and context information. By replacing important
words with rare synonyms, we created  downstream
task datasets 
that are more challenging and
support the evaluation of NLP models on the task of
understanding rare words, a capability that human speakers have.
On all of these
datasets, \textsc{Bertram} improves over standard BERT and RoBERTa,
demonstrating the usefulness of our  method.

Our analysis showed that \textsc{Bertram}
is beneficial not only for rare words (our main target in this paper), but also for frequent words.
In future work, we want to investigate \textsc{Bertram}'s
potential benefits for  such frequent words. Furthermore, it would be interesting to explore more complex ways of incorporating surface-form information -- e.g., by using a character-level CNN similar to the one of \citet{kim2016character} -- to balance out the potency of \textsc{Bertram}'s form and context parts.

\section*{Acknowledgments}
This work was funded by the European Research Council (ERC \#740516).
We would like to thank the anonymous reviewers
for their helpful comments.

\bibliography{literatur}
\bibliographystyle{acl_natbib}

\clearpage
\appendix

\section{Training Details}
\label{appendix-training-parameters}

Our implementation of \textsc{Bertram} is based on PyTorch \citep{paszke2017automatic} and the Transformers library \citep{wolf2019transformers}. To obtain target embeddings for frequent multi-token words (i.e., words that occur at least 100 times in WWC+BookCorpus) during training, we use one-token approximation (OTA) \citep{schick2019ota}. For RoBERTa$_\text{large}$, we found increasing the number of iterations per word from $4{,}000$ to $8{,}000$ to produce better OTA embeddings using the same evaluation setup as \citet{schick2019ota}. For all stages of training, we use Adam \citep{kingma2014adam} as optimizer.

\paragraph{Context-Only Training.} During the first stage of our training process, we train \textsc{Bertram} with a maximum sequence length of 96 and a batch size of 48 contexts for BERT$_\text{base}$ and 24 contexts for RoBERTa$_\text{large}$. These parameters are chosen such that a batch fits on a single Nvidia GeForce GTX 1080Ti. Each context in a batch is mapped to a word $w$ from the set of training words, and each batch contains at least 4 and at most 32 contexts per word. For BERT$_\text{base}$ and RoBERTa$_\text{large}$, we pretrain the context part for 5 and 3 epochs, respectively. We use a maximum learning rate of $5 \cdot 10^{-5}$ and perform linear warmup for the first 10\% of training examples, after which the learning rate is linearly decayed.

\paragraph{Form-Only Training.}

In the second stage of our training process, we use the same parameters as \citet{schick2019ota}, as our form-only model is the very same as theirs. That is, we use a learning rate of $0.01$, a batch size of $64$ words and we apply $n$-gram dropout with a probability of 10\%. We pretrain the form-only part for 20 epochs.

\paragraph{Combined Training.} For the final stage, we use the same training configuration as for context-only training, but we keep $n$-gram dropout from the form-only stage. We perform combined training for 3 epochs. For \textsc{add}, when using RoBERTa as an underlying language model, we do not just prepad the input with the surface-form embedding followed by a colon, but additionally wrap the surface-form embedding in double quotes. That is, we prepad the input with $e_{"}, v_{(w,C)}^\text{form}, e_{"}, e_:$. We found this to perform slightly better in preliminary experiments with some toy examples.

\section{Evaluation Details}

\paragraph{WNLaMPro} In order to ensure comparability with results of \citet{schick2019ota}, we use only WWC to obtain contexts for WNLaMPro keywords.

\paragraph{Rarified Datasets} To obtain rarified instances of MNLI, AG's News and DBPedia, we train BERT$_\text{base}$ and RoBERTa$_\text{large}$ on each task's training set for 3 epochs. We use a batch size of 32, a maximum sequence length of 128 and a weight decay factor of $0.01$. For BERT, we perform linear warmup for the first 10\% of training examples and use a maximum learning rate of $5\cdot 10^{-5}$. After reaching its peak value, the learning rate is linearly decayed. For RoBERTa, we found training to be unstable with these parameters, so we chose a lower learning rate of $1 \cdot 10^{-5}$ and performed linear warmup for the first $10{,}000$ training steps.

To obtain results for our baselines on the rarified datasets, we use the original Mimick implementation of \citet{pinter2017mimicking}, the A La Carte implementation of \citet{khodak2018carte} and the Attentive Mimicking implementation of \citet{schick2019attentive} with their default hyperparameter settings. As A La Carte can only be used for words with at least one context, we keep the original BERT embeddings whenever no such context is available.

While using \textsc{Bertram} allows us to completely remove the original BERT embeddings for all rare words and still obtain improvements in accuracy on all three rarified downstream tasks, the same is not true for RoBERTa, where removing the original sequence of subword token embeddings for a given rare word (i.e., not using the \textsc{slash} variant) hurts performance with accuracy dropping by 5.6, 7.4 and 2.1 points for MNLI, AG's News and DBPedia, respectively. We believe this to be due to the vast amount of additional contexts for rare words in RoBERTa's training set that are not available to \textsc{Bertram}.

\end{document}